\documentclass[runningheads]{llncs}
\usepackage{graphicx}
\usepackage{hyperref}
\usepackage{support-caption}
\usepackage{subcaption}

\usepackage{graphicx} 
\usepackage{mathtools}
\usepackage{amssymb}
\usepackage{multirow}
\begin{document}
\title{Explaining Relation Classification Models with Semantic Extents}
%
%
\author{Lars Klöser\inst{1}\orcidID{0000-0002-0931-8977} \and
Andre Büsgen\inst{1}\orcidID{0000-0002-8298-2567} \and
Philipp Kohl\inst{1}\orcidID{0000-0002-5972-8413} \and
Bodo Kraft\inst{1} \and
Albert Zündorf\inst{2}}
\authorrunning{L. Klöser et al.}
%
\institute{
Aachen University of Applied Sciences, 52066 Aachen, Germany 
\email{\{kloeser,buesgen,p.kohl,kraft\}@fh-aachen.de}
\and
University of Kassel, 34109 Kassel, Germany
\email{zuendorf@uni-kassel.de}\\
}
\maketitle              
\begin{abstract}
In recent years, the development of large pretrained language models, such as BERT and GPT, significantly improved information extraction systems on various tasks, including relation classification. State-of-the-art systems are highly accurate on scientific benchmarks. A lack of explainability is currently a complicating factor in many real-world applications. Comprehensible systems are necessary to prevent biased, counterintuitive, or harmful decisions.

We introduce semantic extents, a concept to analyze decision patterns for the relation classification task. Semantic extents are the most influential parts of texts concerning classification decisions. Our definition allows similar procedures to determine semantic extents for humans and models. We provide an annotation tool and a software framework to determine semantic extents for humans and models conveniently and reproducibly. Comparing both reveals that models tend to learn shortcut patterns from data. These patterns are hard to detect with current interpretability methods, such as input reductions. Our approach can help detect and eliminate spurious decision patterns during model development. Semantic extents can increase the reliability and security of natural language processing systems. Semantic extents are an essential step in enabling applications in critical areas like healthcare or finance. Moreover, our work opens new research directions for developing methods to explain deep learning models.

\end{abstract}
 
\section{Introduction}
\label{introduction}
Digitalizing most areas of our lives lead to constantly growing amounts of available information. With digital transformation comes the need for economic and social structures to adapt. This adaption includes the structuring and networking of information for automated processing. Natural language is a significant source of unstructured information. While comfortable and efficient for humans, processing natural language remains challenging for computers. Structuring and networking the amount of text created daily in sources like online media or social networks requires incredible manual effort. 

The field of \textit{information extraction} (IE) is a subfield of \textit{natural language processing} (NLP) and investigates automated methods to transform natural language into structured data. It has numerous promising application areas; examples are the extraction of information from online and social media \cite{kloserMultiAttributeRelationExtraction2021}, the automatic creation of knowledge graphs \cite{dsouzaSemEval2021Task112021}, and the detailed analysis of behaviors in social networks \cite{busgenExploratoryAnalysisChatbased2022}.

Modern deep-learning-based NLP systems show superior performance on various benchmarks compared to humans. For example, on average deep learning systems like \cite{zhongEfficientLanguageModel2022} score higher than humans on the \textit{SuperGLUE}\footnote{\url{https://super.gluebenchmark.com/}} \cite{wangSuperGLUEStickierBenchmark2020} benchmark. \cite{schlangenTargetingBenchmarkMethodology2020} discusses that such results do not directly imply superior language skills. A primary reason is that the decision-making processes are not intuitively comprehensible. NLP systems are optimized on datasets and therefore run the risk of learning counterintuitive patterns from the data \cite{fengPathologiesNeuralModels2018,gardnerEvaluatingModelsLocal2020}. This research focuses on \textit{relation classification}, a central task in IE, and aims to compare the central decision patterns of human and deep-learning-based task-solving.

\begin{figure}
    \centering
    \includegraphics[scale=.55]{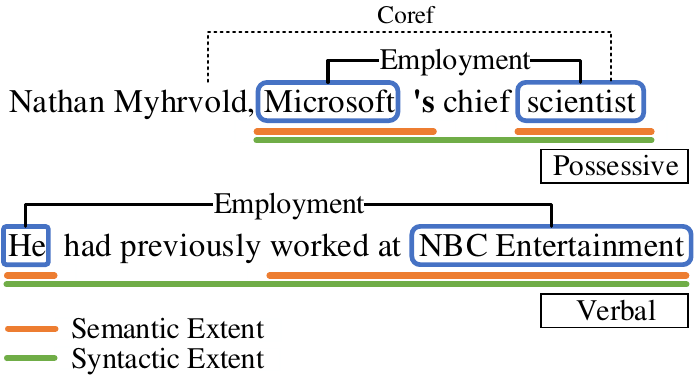}
    \caption{Two relation mentions from the ACE 05 corpus. The first example shows a possessive formulation. The second example shows a verbal formulation. In the first example, \textit{scientist} is preferable to \textit{Nathan Myhrvold}. \textit{scientist} is part of a predefined class of formulations (possessive), while \textit{Nathan Myhrvold} is not.}
    \label{fig:semantic_extent}
\end{figure}

In more detail, relation classification describes classifying the relation for a given pair of entity mentions, the relation candidate, in a text. \autoref{fig:semantic_extent} shows two examples for the task. We assume that one of the possible relations applies to the relation candidate. A real-world application requires additional steps like detecting entity mentions or numerating relation candidates.  We analyze a deep-learning NLP model on the ACE 05 dataset \cite{doddingtonAutomaticContentExtraction2004}. Our primary focus is the comparison between human and deep-learning decision patterns. We introduce the concept of \textit{semantic extents} to identify human decision patterns. 

\cite{doddingtonAutomaticContentExtraction2004} introduced the related concept of \textit{syntactic classes} and \textit{syntactic extents}. Natural language offers various syntactical ways to formulate relationships between mentions of certain entities in text. Each involves different syntactical aspects of a sentence. For example, as shown in \autoref{fig:semantic_extent}, a relationship can be expressed through a possessive formulation or a verbal construction. In these examples, \textit{possessive} and \textit{verbal} are the relation's syntactic classes. The syntactic extent contains both arguments and the text between them. If necessary for the syntactic construction implied by the syntactic class, it contains additional text outside the arguments. The ACE corpus contains only relations that fit into one of the given syntactic classes. 

Syntactic extents neglect the fact that the semantics of a decision can strongly influence decision patterns. Semantic extents are minimum necessary and sufficient text passages for classification decisions concerning a specific \textit{decider}. Humans, as well as NLP models, can be deciders. We introduce a tool for practical human annotation of semantic extents. The semantic extents for humans and models show significant differences. Our results indicate that human decisions are more context-dependent than model decisions. Models tend to decide based on the arguments and ignore their context. We provide an analysis based on adversarial samples to support this assumption. Finally, we compare semantic extents to other existing explainable \textit{artificial intelligence} (AI) techniques. The results of this study have the potential to reveal incomprehensible and biased decision patterns early on in the development process, thus improving the quality and reliability of IE systems.

We provide reproducible research. All source code necessary to reproduce this paper's results are available via GitHub \footnote{\url{https://github.com/MSLars/semantic_extents} \url{https://github.com/MSLars/ace_preprocessing}}. The paper aims to investigate and compare the decision patterns of human and deep-learning-based models concerning relation classification. Specifically, we analyze the concept of semantic extents and compare them concerning human and model decisions. The paper also explores existing explainable AI techniques for relation classification. The primary research question can be summarized as follows:

\begin{center}
    \textit{
   What are the decision patterns of human and deep-learning-based models when solving the task of relation classification, and how do they compare?  
   }
\end{center}

Our main contributions are:

\begin{enumerate}
    \item the introduction and analysis of semantic extents for relation classification,
    \item the first analysis of explainable AI techniques like input reductions for relation classification, and
    \item a novel preprocessing pipeline based on a state-of-the-art python tool stack for the ACE 05 dataset.
\end{enumerate}

\section{Related Work} 
\label{sec:related_work}

This section surveys different approaches and techniques proposed to interpret and explain the decisions made by NLP models and discusses their strengths and limitations.

The best information extraction approaches on scientific benchmarks finetune large pretrained language models like BERT \cite{devlinBERTPretrainingDeep2019} and LUKE \cite{yamadaLUKEDeepContextualized2020}. Various studies try to understand the internal processes that cause superior performance on various NLP benchmarks.

One direction of research focuses on analyzing how models represent specific meanings. Besides the superior performance on benchmarks, pretrained text representations may be the source of harmful model behaviors. \cite{bolukbasiManComputerProgrammer2016} shows that static word embeddings, such as Word2Vec, tend to be biased toward traditional gender roles. Approaches, such as \cite{nissimFairBetterSensational2020,zhangMitigatingUnwantedBiases2018,zhaoMenAlsoShopping2017}, tried to remove these biases. \cite{brunetUnderstandingOriginsBias2018} investigate the origins of biases in these vector representations.
Similarly, \cite{liDetectingGenderBias2021} showed that pretrained transformer language models suffer from similar biases. These results show that public data sources are biased. This property could carry over to finetuned models. 

\cite{mengLocatingEditingFactual2023} focused on the internal knowledge representation of transformer models and showed that they could change specific factual associations learned during pretraining. For example, they internalized the fact that \textit{the Eifel tower is in Rome} in a model.

Another direction of research focuses on the language capabilities of pretrained language models. \cite{clarkWhatDoesBERT2019} showed that BERT's attention weights correlate with various human-interpretable patterns. They concluded that BERT models learn such patterns during pretraining. \textit{Probing} is a more unified framework for accessing the language capabilities of pretrained language models. \cite{wuPerturbedMaskingParameterfree2020} provided methods for showing that language models learn specific language capabilities. \cite{tenneyBERTRediscoversClassical2019} showed that latent patterns related to more complex language capabilities tend to occur in deeper network layers. Their findings suggest that pretrained transformer networks learn a latent version of a traditional NLP pipeline.

The previously mentioned approaches investigate the internal processes of large pretrained language models. Another research direction is analyzing how finetuned versions solve language tasks in a supervised setting.
\cite{mccoyRightWrongReasons2019} created diagnostic test sets and showed that models learn heuristics from current datasets for specific tasks. Models might learn task-specific heuristics instead of developing language capabilities. \cite{ribeiroAccuracyBehavioralTesting2020} provided a software framework for similar tests and revealed that many, even commercial systems, fail when confronted with so-called diagnostic test sets or adversarial samples.

A different way to look at the problem of model interpretability is the analysis of single predictions. \cite{simonyanDeepConvolutionalNetworks2013} introduced gradient-based \textit{saliency maps}. \cite{sundararajanAxiomaticAttributionDeep2017,smilkovSmoothGradRemovingNoise2017} adopted the concept for NLP. Each input token is assigned a saliency score representing the influence on a model decision. \textit{Input reductions} remove supposedly uninfluential tokens. They interpret the result as a semantic core regarding a decision. This concept is similar to semantic extents, except that we extend and they reduce input tokens.
Similarly, \cite{ebrahimiHotFlipWhiteBoxAdversarial2018} rearrange the input. If perturbation in some area does not influence a model's decision, the area is less influential. Besides gradient information from backpropagation, researchers analyze the model components' weights directly. \cite{clarkWhatDoesBERT2019} visualizes the transformer's attention weights. Their results indicate that some attention heads' weights follow sequential, syntactical, or other patterns.

\cite{shahbaziRelationExtractionExplanation2020} introduce an explainability approach for relation extraction with distant supervision. They use a knowledge base to collect text samples, some expressing the relationship between entities and others not. The model then predicts the relationship between these samples with methods that assess the model's relevance for each sample.

Meanwhile, \cite{ayatsTwoStepApproachExplainable2022} brings us a new development in the field with their interpretable relation extraction system. By utilizing deep learning methods to extract text features and combining them with interpretable machine learning algorithms such as decision trees, they aim to reveal the latent patterns learned by state-of-the-art transformer models when tasked with determining sentence-level relationships between entities.

The field of explainable NLP covers various aspects of state-of-the-art models. We are a long way from fully understanding decisions or even influencing them in a controlled manner. Various practical applications, especially in sensitive areas, require further developments in this field.

\section{Dataset}
\label{sec:dataset}

The ACE\footnote{Short for Automated Content Extraction} 05 dataset is a widely used benchmark dataset for NLP tasks such as entity, event, and relation extraction. It was a research initiative by the United States government to develop tools for automatically extracting relevant information from unstructured text. The resources are not freely available.


The dataset consists of news articles from English-language newswires from 2003 to 2005. This paper analyzes explainability approaches for relation classification models using the ACE 05 dataset. This dataset includes annotations for various entities, relations, and events in 599 documents. Entities, relations, and events may have various sentence-level mentions. We focus on entities and relations and leave out the event-related annotations. 


Our analysis focuses on the sentence-level relation mentions annotated in the ACE 05 dataset. The annotations have additional attributes, such as tense and syntactic classes, often ignored by common preprocessing approaches for event \cite{zhangJointEntityEvent2019} and relation \cite{liIncrementalJointExtraction2014} extraction. We publish a preprocessing pipeline on top of a modern Python technology stack. Our preprocessing approach makes this meta-information easily accessible to other researchers. We refer to the original annotation guidelines for further information about the relation types and syntactic classes \footnote{\url{https://www.ldc.upenn.edu/sites/www.ldc.upenn.edu/files/english-relations-guidelines-v5.8.3.pdf}}. Our preprocessing pipeline is available via a GitHub repository.


\subsection{Preprocessing}
\label{sec:preprocessing}

Our preprocessing pipeline includes all data sources provided in the ACE 05 corpus. In contrast, \cite{liIncrementalJointExtraction2014} excludes all Usenet or forum discussions and phone conversation transcriptions. However, real-world NLP systems might encounter informal data from social media and comparable sources. We see no reason to exclude these relations. Our train-dev-test split expands \cite{liIncrementalJointExtraction2014}. We extended each split with random samples from the previously excluded data sources and kept the ratio between all three splits similar.

The currently most widespread pipelines use a combination of Java and Python implementations. The sentence segmentation in \cite{liIncrementalJointExtraction2014} uses a Standford NLP version from 2015. Our implementation uses the Python programming language and the current version of spaCy\footnote{\url{https://spacy.io/}, version 3.5}. All source codes are open for further community development since the technologies used are common in the NLP community.


We offer a representation of the data structures from the original corpus in a modern Python environment. In contrast to existing approaches, we can access cross-sentence coreference information. Users can access the coreference information between \textit{Entities}, \textit{Relations}, and \textit{Events}. Our preprocessing enables researchers to use the ACE 05 corpus for further tasks such as document-level relation extraction or coreference resolution.

We use spaCy's out-of-the-box components for tokenization and sentence segmentation. Each \textit{relation mention} forms a relation classification sample if both arguments are in one detected sentence. Additionally, one sentence may occur in multiple relation classification samples if it contains multiple \textit{relation mentions}.

\subsection{Task Definition}
\label{sec:task_definition}

We define the relation classification task as classifying the relationship between two argument spans in a given sentence. Given an input sentence $s = x_1, ..., x_n$ that consists of $n$ input tokens. We refer to sentences in the sense of ordinary English grammar rules. In relation classification, two argument spans, $a_1 = x_i, ..., x_j$ and $a_2 = x_k, ..., x_l$ with $1 \leq i < j < k < l \leq n$, are defined as relation arguments. The model's task is to predict a label from a set of possible labels $l \in R$, where $R$ is the set of all possible relation labels. It is important to note that the model only considers samples where one of the relations in $R$ applies to the pair of arguments.

Our definition of relation classification differs from relation extraction. Relation extraction typically involves identifying entity mentions in the text and then analyzing their syntactic and semantic relationships. Relation classification, conversely, involves assigning predefined relationship types to given pairs of entities in text. In this task, we provide the relation's arguments, and the goal is to classify each pair of entities into the correct predefined relationship category.

\subsection{Dataset Statistics}
\label{sec:dataset_statistics}

\begin{figure}[ht]
    \centering
    \begin{subfigure}[b]{.8\textwidth}
    \includegraphics[scale=.5]{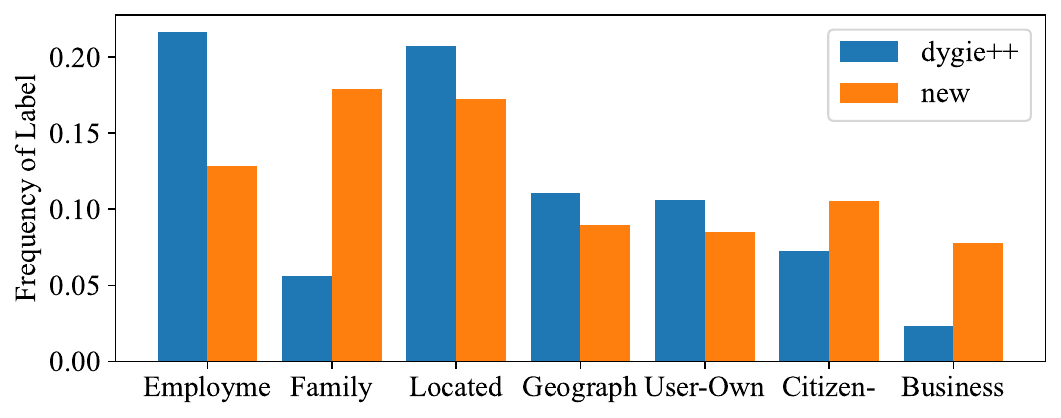}
    \caption{Most frequent relation labels.}
    \end{subfigure}

    \begin{subfigure}[b]{.8\textwidth}
    \includegraphics[scale=.5]{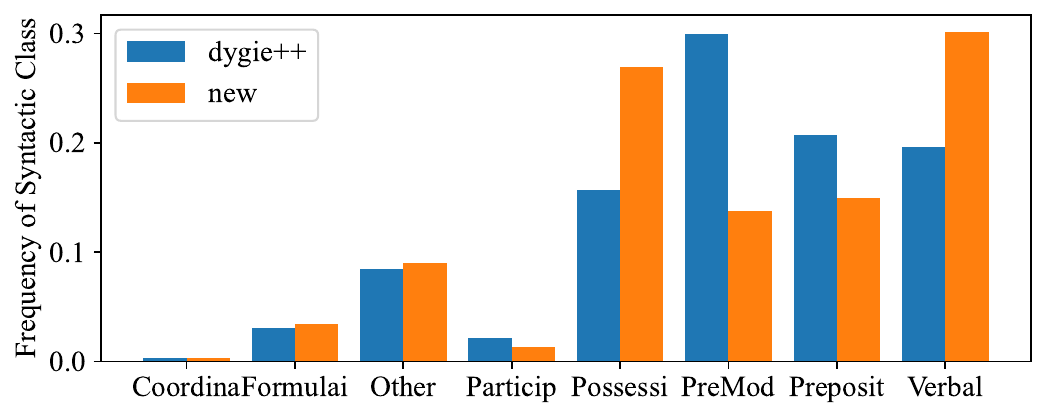}
    \caption{Most frequent syntactic classes.}
    \end{subfigure}
    \caption{The histograms show the semantic and syntactic properties of different parts of the ACE 05 dataset. \textit{dygie++} refers to all samples in \cite{liIncrementalJointExtraction2014,wadden-etal-2019-entity}. The name originates from the implementation we used. \textit{new} refers to the excluded samples.}
    \label{fig:dataset_stats}
\end{figure}


By adding Usenet or forum discussions and phone conversation transcriptions as data sources, we increase the semantical and syntactical scope of the dataset. \autoref{fig:dataset_stats} shows the distribution of relation labels and syntactic classes. While the more informal sources contain more \textit{Family} relations, the other sources contain more business-related \textit{Employer} relations. Similarly, other sources contain less \textit{Possessive} and \textit{Verbal} formulations. The results indicate that adding the new sources extends the semantic and syntactic scope of the dataset since added samples have different characteristics.

\section{Semantic Extent}
\label{sec:decision_areas}

To answer the main research question, we formalize the concept of \textit{decision patterns}. Decision patterns are text areas contributing significantly to decisions in the context of an NLP task. We refer to these significantly influential text areas as the \textit{semantic extent}. The semantic extent contains the relation's arguments for the relation classification task.

We need four additional concepts to specify the scope of semantic extents. Given an input sentence $s = x_1, ..., x_n$, each ordered subset $s_{cand} \subset s$ that contains the arguments can be interpreted as a \textit{semantic extent candidate}. A \textit{candidate extension}, $s_{ext} = s_{cand} \cup x_i$ for some $x_i \notin s_{cand}$, is the addition of one token. 

An extension is \textit{consistent} if the added token does not substantially reduce the influence of other tokens of the candidate. If no extension of a candidate that causes the classification decision to change exists, then the candidate is \textit{significant} and referred to as a semantic extent.

\begin{figure}
    \centering
    \includegraphics[scale=.45]{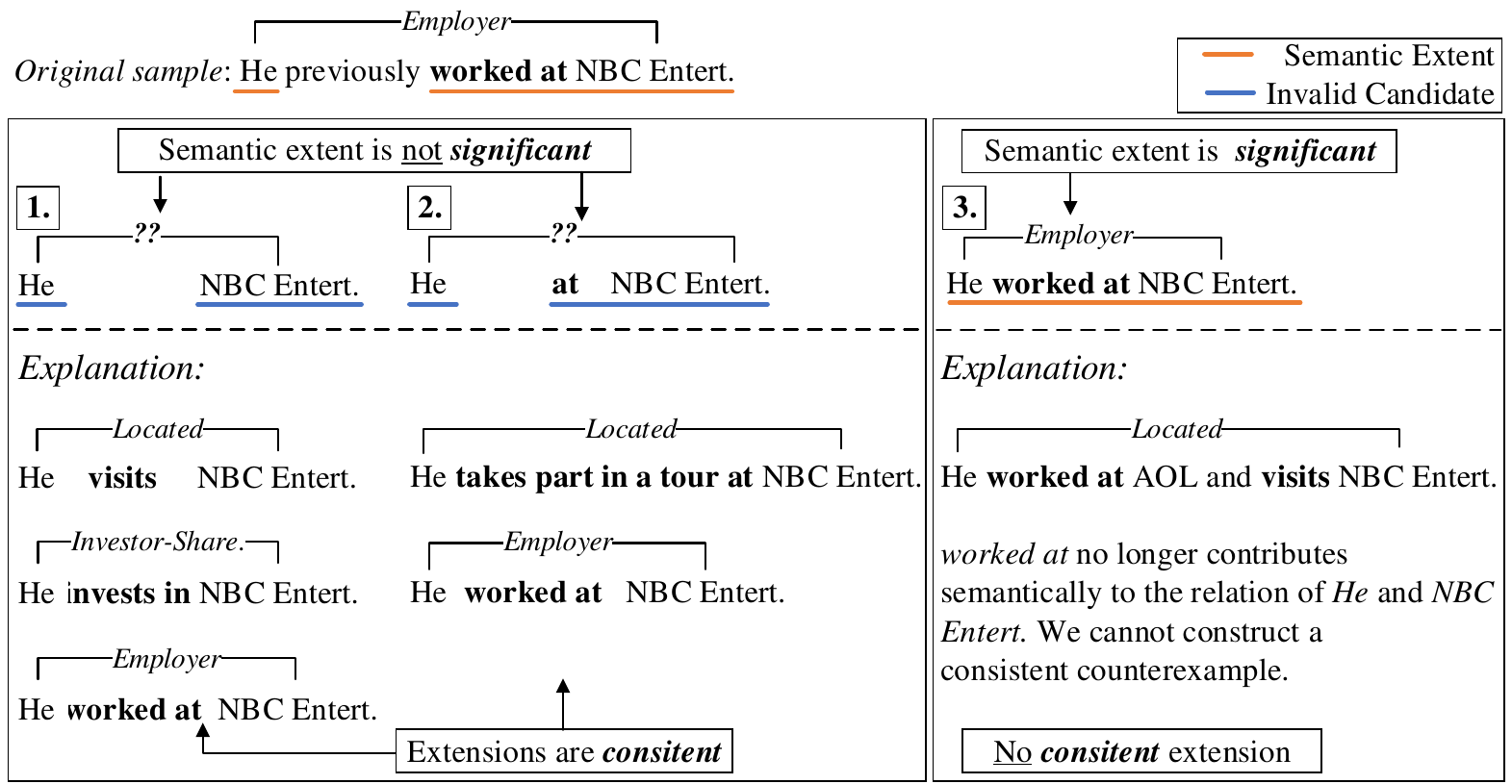}
    \caption{Example determination of the semantic extent The top row contains suggestions for the semantic extent. Each column contains extensions that allow a different interpretation. The last column contains the valid semantic extent for the sample \textit{He had previously worked at NBC Entertainment}. The top row shows different semantic extent candidates. Each column list possible extensions that change the relation label.}
    \label{fig:sem_extent_definition_example}
\end{figure}

\autoref{fig:sem_extent_definition_example} shows how to apply the previous definitions to determine a semantic extent. We aim to explain why the sentence \textit{He had previously worked at NBC Entertainment} expresses an \textit{Employer} relation between \textit{He} and \textit{NBC Entertainment}.

Are the arguments themselves a valid semantic extent? Our first candidate is \textit{He NBC Entert.} One consistent extension could be \textit{He visits NBC Entertainment}. It suggests a \textit{Located} relation. Therefore, \textit{He NBC Entertainment} is not a significant candidate nor a semantic extent. To create the next semantic extent candidate, we add \textit{at}. As the second column in \autoref{fig:sem_extent_definition_example} suggests, we can still create consistent extensions that cause a different classification decision. Finally, \textit{He works at NBC Entertainment} does not allow consistent extensions that change the classification label. \textit{He works at NBC Entertainment} is the semantic extent.

However, extensions that change the classification decision are still possible. For example, \textit{He works at AOL and visits NBC Entertainment}. \textit{works at}, as part of the semantic extent, does not influence the classification decision. \textit{visits} is this extension's most significant context word. We do not consider this a consistent extension and assume that no consistent extension exists.

Generally, it is impossible to prove the absence of any consistent extension changing the relation label. Nevertheless, we argue that the concepts of \textit{interpretability} and \textit{explainability} suffer similar conceptual challenges. We apply a convenient extension rule: if an average language user cannot specify a convenient extension, we assume such does not exist. Our evaluation uses the additional concept of \textit{semantic classes}. Semantic classes group different kinds of semantic candidate extensions. The following section motivates and describes the concept in detail.

\subsection{Semantic Extent for Humans}
\label{sec:semantic_extent_for_humans}

This section proposes a framework and the corresponding tooling for human semantic extent annotations. We explain and justify the most relevant design decisions. In general, many other setups and configurations are possible. A comparative study is out of the scope of this research.

A possible approach would be to show annotators the complete text and task them to mark the semantic extent directly. With this approach, annotators manually reduce the input presented to them. We assume that annotators would annotate to small semantic extents because the additional information biases them. We, therefore, define an extensive alternative to such reductive approaches.

\begin{figure}[ht]
    \centering
    \includegraphics[scale=.25]{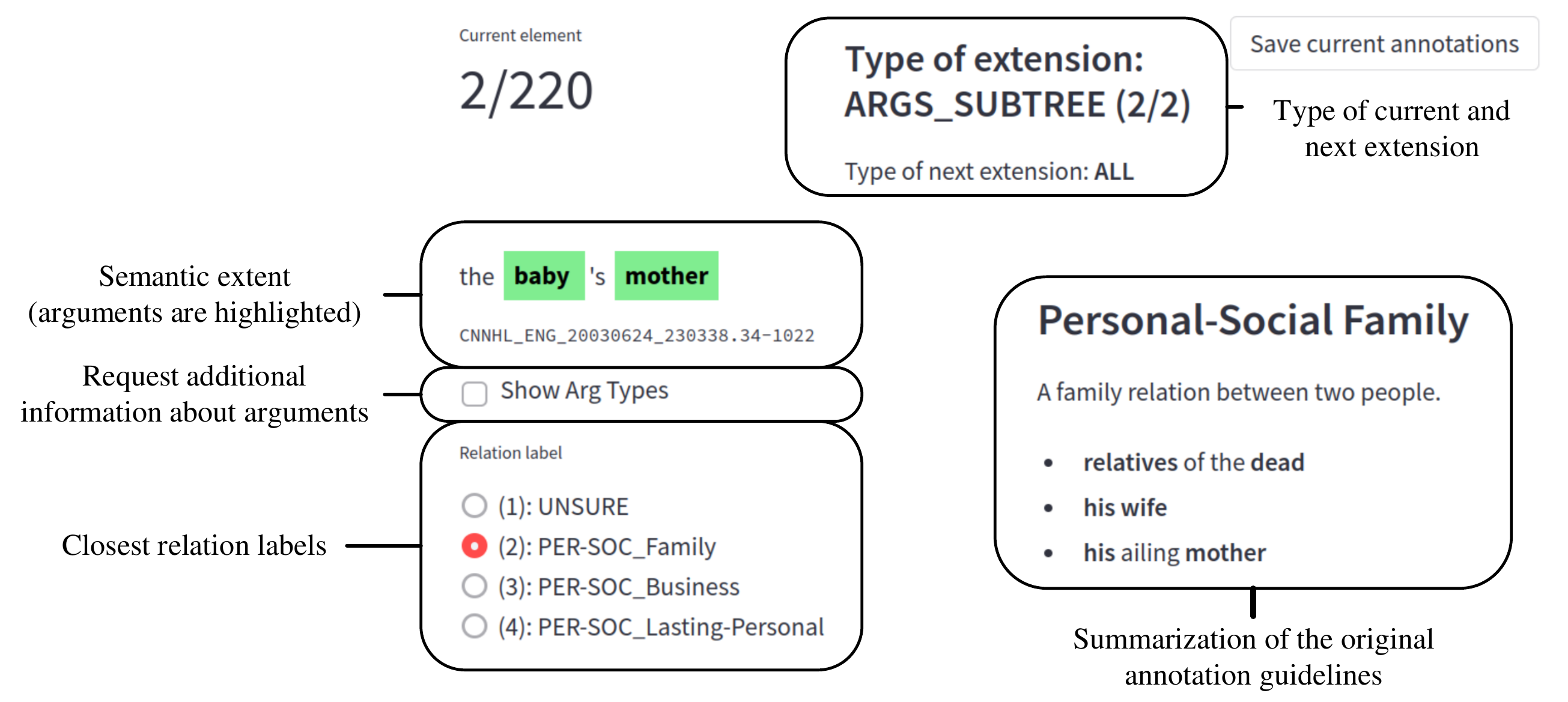}
    \caption{Screenshot of our annotation tool's central area. Annotators use the keyboard for label selection, navigation between samples, and extension of the semantic extent.}
    \label{fig:annotation_tool}
\end{figure}

We start with confronting annotators with only the argument text spans. If they are sure to determine the relation label, they can annotate the relation and jump to the following sample. If not, they can request further tokens. This procedure continues until they classify a sample or reject a classification decision. \autoref{fig:annotation_tool} shows a screenshot of our annotation tool. 

The extension steps in \autoref{fig:sem_extent_definition_example} are examples of such an iterative extension. At first, an annotator sees the arguments \textit{He} and \textit{NBC Entertainment}. The first extension reveals an \textit{at}, and the final extension reveals the verb \textit{works}. We implemented a heuristic for the priority of words in this process. 

\begin{figure}[ht]
    \centering
    \includegraphics[scale=.45]{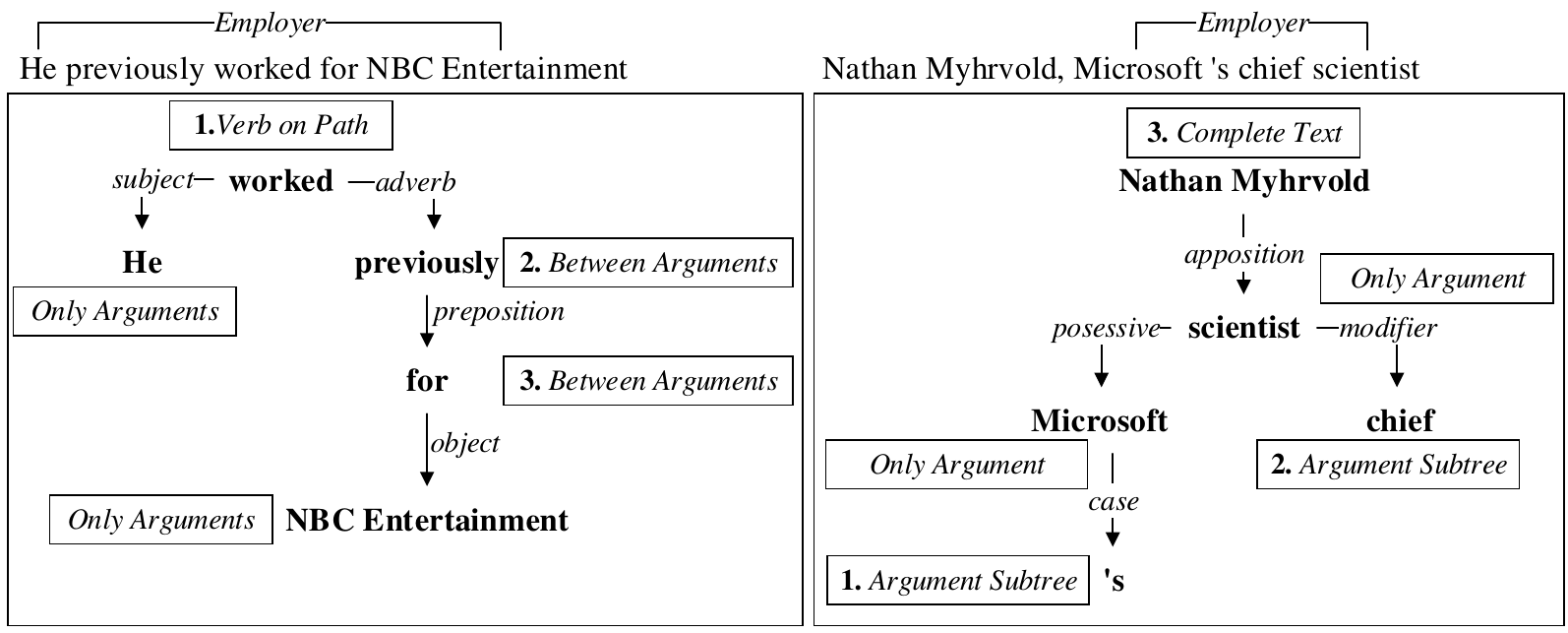}
    \caption{Example of semantic extension types in combination with the dependency parse tree for two samples. The indices show the priority index during the expanding annotation process.}
    \label{fig:semantic_extent_priority}
\end{figure}

Our priority can be summarized as follows. The initial semantic extent candidate contains only arguments (OA). The second stage contains the syntactic subtrees of the arguments (AS). The third stage shows the verbs on the syntactic path between the arguments (VOP). Afterward, we show all tokens between the arguments (BA). Finally, we show the full semantic extent annotated in the ACE 05 corpus (E) and all remaining tokens (A) if necessary. Some of these stages may contain multiple or even none tokens. \autoref{fig:semantic_extent_priority} illustrates our prioritization for two examples. We refer to the category of the final extension as \textit{semantic class}. We use this class extensively during the evaluation in \autoref{sec:evaluation}.

We implemented some simplifications to increase practicability and reduce the necessary human annotation effort.

\begin{enumerate}
    \item Annotators had to check the annotation guidelines to select possible relation candidates. The dataset contains 18 different relation labels. Adequate familiarization with the annotation scheme is too time-consuming. We implemented a label preselection. Annotators choose between the three most likely labels (concerning our deep learning model). The preselection reduces the need to know every detail of the annotation guidelines but makes the semantic challenge comparable. Additionally, we provide a summary of the annotation guidelines.
    \item Many expressions needed to be looked up by the annotators. For example, the ACE 05 corpus contains many sources about the war in Iraq. Names of organizations or locations were unknown. We offered annotators the possibility to check the fine-grained argument entity types. This possibility usually avoids additional look-ups and speeds up the annotation process.
    \item We implement the previously mentioned priority heuristic about which tokens are candidates for extensions of the semantic extent. In general, this is necessary for our extensive setting. However, since annotators cannot annotate custom semantic extents, unnecessary tokens might be part of the semantic extent. We use the extension categories for our analysis. For each annotation, we can interpret these semantic classes as statements like \textit{As annotator, I need the verb to classify the sample}.
\end{enumerate}

We had to distinguish between practicability and the danger of receiving biased annotations. However, we assume that our adoptions did not influence or bias the annotations in a significant way. The source code for the annotation process is available in our GitHub repository, so the results are reproducible with other design decisions.

\subsection{Semantic Extent for Models}
\label{sec:model_setup}

\begin{figure}
    \centering
    \includegraphics[scale=.40]{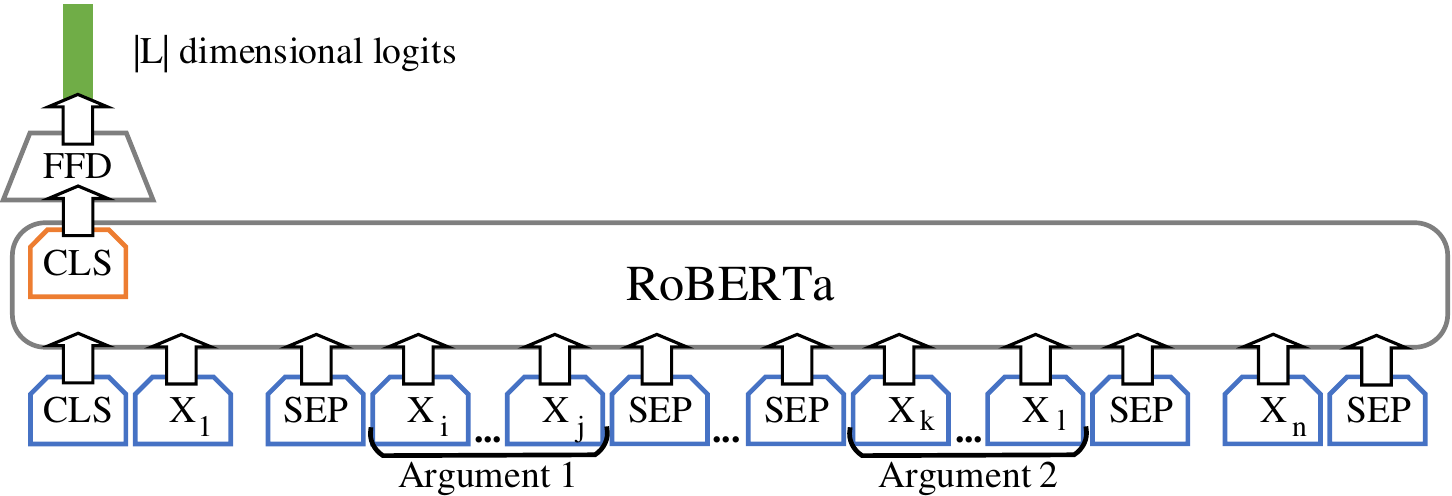}
    \caption{Illustration of our deep learning model architecture. We special \textit{SEP} (separator) tokens to indicate the argument positions.}
    \label{fig:model_architecture}
\end{figure}

This section describes how we determine semantic extents for deep learning models. Finetuned transformer networks are fundamental for most state-of-the-art results in NLP approaches. We define a custom input format for relation classification and finetune a pretrained transformer model, \textit{RoBERTa}, for relation classification. \autoref{fig:model_architecture} illustrates our deep learning setup. 

We tokenize the inputs and use the model-specific \textit{SEP} tokens to indicate the argument positions. To receive \textit{logits}, we project the final representation of the \textit{CLS} token to the dimension of the set of possible relation labels. Logits are a vector representation $h \in \mathbb{R}^{|L|}$  assigning each label a real-valued score. We apply the softmax function to interpret these scores as probabilities
$$
P(l) = \frac{\exp\left(h_l\right)}{\sum_{k \in L}\exp\left(h_k\right)}, \;\forall \, l \in L
$$
and calculate the cross-entropy loss to optimize the model.

Our approach requires a single forward pass to a transformer network. The computational complexity is comparable to many current state-of-the-art approaches. 

As described in the previous sections, the semantic extent is a mixture of reductive and expanding input variations. We use this as a guide and define two methods for determining text areas interpretable as semantic extent for deep learning models.

\subsection{Expanding Semantic Extent}
\label{sec:expanding_extent}

Compared to human annotation, the expanding semantic extent is determined similarly (\autoref{sec:human_annotation}). At first, we calculate the model prediction $l_{all}$ on a complete sentence. Then we create semantic extent candidates. The initial candidate contains only the arguments. We expand the extent by adding words concerning the priority described in \autoref{sec:human_annotation} and compute  a model prediction $l_{reduced}$ on the candidate. If $l_{all} = l_{reduced}$ and $P(l_{reduced}) > \theta$ for some threshold $\theta$, we interpret the current candidate as semantic extent. We interpret the category of the final extension as the semantic class.

\subsection{Reductive Semantic Extent}
\label{sec:reductive_extent}

We apply gradient-based \textit{input reductions} \cite{fengPathologiesNeuralModels2018} to determine a reductive semantic extent. Like \autoref{sec:expanding_extent}, input reductions are a minimal subset of the input with an unchanged prediction compared to the complete input. The main difference is that input reductions remove a maximal number of tokens. Our implementation uses the input reduction code in \cite{wallaceAllenNLPInterpretFramework2019}.

We apply a beam search to reduce the input as far as possible while keeping the model prediction constant. We compute an approximation for the token influence at each reduction step based on gradient information. Let $p$ be the probability distribution predicted by a model and $l_{pred}$ be the predicted label. In the next step we compute a loss score $L_{inter}\left(l_{pred}, p \right)$. We use the gradient $\frac{\partial L_{inter}}{\partial x_{i}}$, for each $x_{i}$ in the current input, to approximate the influence of each token on the model prediction. During the beam search, we try to exclude the least influential tokens.

Reductive semantic extents may be sentence fragments without grammatical structure and unintuitive for humans. If the essential structure deviates too much from the data set under consideration, it is questionable whether we can interpret the reductions as decision patterns. Expanding semantic extents try to get around this by creating sentence fragments that are as intuitively meaningful as possible through prioritization. Following \autoref{sec:evaluation} investigates the characteristics of both approaches in detail.

\section{Evaluation}
\label{sec:evaluation}

In the previous section, we formalized decision patterns for the relation classification task by introducing \textit{semantic extents}. This section compares semantic extents for humans and deep learning models. We reveal similar and dissimilar patterns related to human and deep learning-based problem-solving. 

\subsection{Human Annotation}
\label{sec:human_annotation}


Preparing for the previously mentioned comparison, we analyze human decision patterns by determining and analyzing semantic extents. Due to human resource constraints, we could not annotate semantic extents on all test samples. The available resources are sufficient to prove an educated guess about human decision patterns. Humans generally require additional context information to the relation arguments to make a classification decision. We assume that \textit{possessive}, \textit{preposition}, \textit{pre-modifier}, \textit{coordination}, \textit{formulaic}, and \textit{participial}\footnote{These refer to the syntactic classes for relation-mention formulations introduced in the ACE 05 annotation guidelines.} formulations require a syntactically local context around arguments. In contrast, \textit{verbal} and \textit{other} formulations require more \textit{sentence-level} information. This section refers to both categories as \textit{local} and \textit{sentence-level} samples. We sampled 220 samples with an equal fraction of local and sentence-level samples to prove our claim. 

\begin{table}
\centering
\caption{A summary showing the human annotations' results on 220 test samples. All F1 scores refer to the relation labels, and \textit{label agreement} (LA) indicates the agreement on these annotations. SC refers to \textit{semantic class}. \textit{fine} refers to all semantic classes, while \textit{coarse} distinguishes only arguments and argument subtree from the others.}
\label{tab:human_f1_and_agreement}

\begin{tabular}{ l c c }
                    & Annotator 1  & Annotator 2      \\ \noalign{\smallskip} \hline \hline \noalign{\smallskip}
 F1 micro           & .85 & .78                 \\
 F1 macro           & .84 & .78                 \\
 LA                 & \multicolumn{2}{c}{.76}   \\
 SC coarse         & \multicolumn{2}{c}{.67}   \\
 SC fine           & \multicolumn{2}{c}{.46}   \\
 SC size           & 5.2 $\pm$ 5.5 & 5.2 $\pm$ 5.9 \\
\end{tabular}

\end{table}

We asked two annotators to annotate semantic extents on these 220 samples using the previously introduced application (\autoref{sec:human_annotation}). \autoref{tab:human_f1_and_agreement} shows central characteristics of the annotations. The first annotator made a more accurate prediction concerning F1 scores, but both annotators showed high agreement on the relation labels. While the agreement on fine-grained semantic classes is moderate, annotators agree more clearly on the coarse classes. Regarding the number of tokens in the semantic extents, both annotators show similar numbers with a comparably high standard deviation. Annotators reported that, in some cases, arguments have large subtrees. Large subtrees require many extensions to see the verb of other sentence-level tokens.

\begin{figure}
    \centering
    \includegraphics[scale=.6]{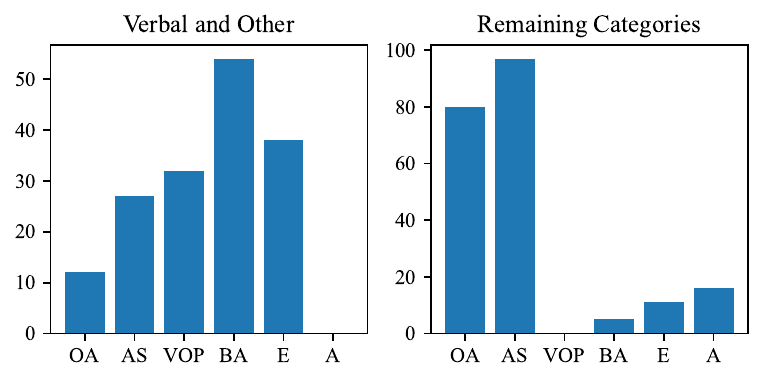}
    \caption{Semantic class of human-annotated semantic extents for different syntactic classes. \textit{Verbal} and \textit{Other} correlate with context-dependent semantic classes. The remaining categories do not.}
    \label{fig:human_semantic_extend}
\end{figure}

\autoref{fig:human_semantic_extend} shows the distribution of semantic classes for local and sentence-level samples. The semantic classes OA and AS indicate that annotators made decisions based on tokens that are children of the relation arguments in a dependency parse tree. The semantic classes VOP, BA, E, and A suggest that annotators needed more sentence-level information for their decision. The syntactic classes \textit{verbal} and \textit{other} also suggest a higher influence of sentence-level contexts. For sentence-level samples, the annotators often need tokens outside the syntactical subtree of either of the arguments. For local samples, the local context suffices for deciding relation labels. Only some cases require additional context information. These additional contexts differ from the previously described sentence-level context. We suppose that annotators had problems classifying these samples and tried to get information from the extensions to avoid excluding examples.

In conclusion, the amount of context information needed for decisions depends on the type of formulation. For the task of relation classification, humans recognize distinguishable patterns. We revealed that humans decide between patterns based on contexts syntactically local to the arguments and contexts that span entire sentences. Most decisions are not possible completely without any context.

\subsection{Model Performance}
\label{sec:model_performance}

The previous section revealed two human decision patterns. Humans either focus on local contexts around the arguments or on sentence-level contexts. This section similarly investigates latent model decision patterns. Besides traditional metrics such as micro and macro F1 scores, we assign decisions to different behavioral classes and investigate the implications on model language capabilities and real-world applications. We use the complete test dataset for the following analyses unless otherwise stated.

\begin{table}[ht]
    \centering
    \caption{Model and human performance on the ACE 05 relation classification corpus. ($^*$) on a subset of 220 samples. ($^{**}$) on the complete test set.}
    \label{tab:model_performance}
    \begin{tabular}{ l  c c c}
                & F1-micro  & F1-macro & \#Parameters  \\ \noalign{\smallskip} \hline \hline \noalign{\smallskip}
     Annotator 1$^*$    &.846        &.843    &        \\
     Annotator 2$^*$    &.779        &.779    &    \\
     RoBERTa base$^{**}$     &.832        &.678   & ~123 million        \\
     RoBERTa large$^{**}$    &.855        &.752   & ~354 million     \\
    \end{tabular}

\end{table}

\autoref{tab:model_performance} presents the transformer models' overall performance compared to humans. We trained two model variants, one with the base version of RoBERTa and the other with the large version. F1-micro scores for annotator 1 and both models are in a similar range\footnote{Since Annotator 1 performs better than annotator 2, we take his scores as a careful estimation of human performance.}. The F1-macro scores reveal that model performances differ across relation labels. An investigation of F1 scores per label shows that RoBERTa-base recognizes neither \textit{Sports Affiliation} nor \textit{Ownership} relation. RoBERTa-large suffers from similar problems in a slightly weaker form. We selected a subset of 220 samples for human annotation. Some effects might be unlikely for underrepresented classes at this sample size. However, concerning the F1-micro score, the results show that transformer-based NLP models solve the sentence-level relation classification task (\autoref{sec:task_definition}) comparably well as humans. 

\subsection{Extensive Semantic Extents}
\label{sec:extensive_semantic_extents}

\begin{figure}[ht]
    \centering
    \includegraphics[scale=.7]{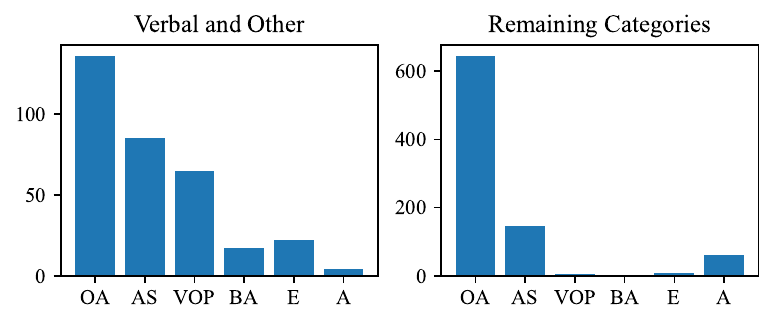}
    \caption{Semantic classes for the expanding semantic extents for the RoBERTa-base model. The two histograms show the semantic classes depending on the syntactic classes.}
    \label{fig:model_semantic_extend}
\end{figure}

The models' semantic extents hint at differences in model and human decision patterns. \autoref{fig:model_semantic_extend} shows the classes of expanding semantic extents (\autoref{sec:expanding_extent}). Confronted with only the arguments, models tend to make confident decisions similar to their decision on complete samples. In comparison, \autoref{fig:human_semantic_extend} indicates that humans cannot make similar decisions based on the arguments. This observation may have different reasons. Models are optimized to reproduce annotations on concrete datasets. These annotations may contain shortcuts and lead models towards heuristics diverging from language capabilities we would expect from humans. Another reason may be that we force models to make confident decisions in common deep-learning setups. Argument texts allow good heuristic solutions, and our setup offers no way to express uncertainty for models. In combination, that may cause overconfident model decisions if certain argument text correlate with particular labels in the dataset.

\begin{table}
    \centering
    \caption{Performance and confidence of the RoBERTa base model on samples with different semantic classes.}
    \label{tab:model_conf_and_performance_semantic_extend}
    \begin{tabular}{l c c c}
                    &Confidence &F1 Micro   &F1-Macro   \\ \noalign{\smallskip} \hline \hline \noalign{\smallskip}
    Complete dataset     &.88$\pm$.16&.82        &.66        \\ \noalign{\smallskip}
    Only Arguments       &.94$\pm$.13&.86        &.69        \\
    Non-Only Arguments   &.82$\pm$.16&.77        &.57        \\ \noalign{\smallskip}
    All tokens in extent     &.74$\pm$.16&.72        &.60        \\
    Not all tokens in extent &.90$\pm$.14&.82        &.65        \\ 
    \end{tabular}

\end{table}

\autoref{tab:model_conf_and_performance_semantic_extend} relates the semantic class with model confidences and the correctness of predictions. We distinguish two categories of decisions. The first summarizes samples for which the model makes the final decision based on arguments alone. Secondly, we focus on samples where the model needs all tokens to make its final decision. Even slight changes in the input change the model prediction for these samples.

Predictions that fall in the first category are significantly more likely to be correct and show higher confidence with less deviation. The results indicate two things. First, in this data set, arguments are often a strong indication of the relation label. Second, models recognize such correlations and use them for decisions. Our experiments indicate that people solve the relation classification task fundamentally differently. This model behavior is beneficial to increase metrics on benchmark datasets. This may result in wrong predictions in real-world applications.

\begin{figure}
    \centering
    \includegraphics[scale=.48]{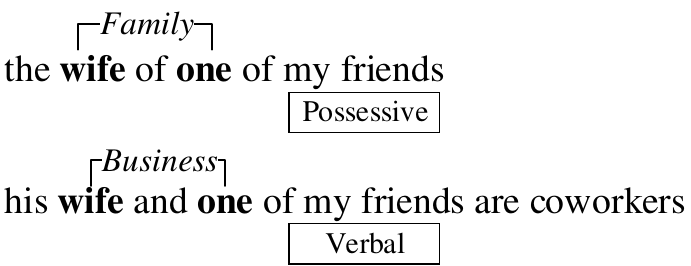}
    \caption{Samples with identical argument text and different relation labels. Correct decisions on both samples require including context information.}
    \label{fig:adversarial_sample}
\end{figure}

\autoref{fig:adversarial_sample} shows examples giving intuition that models may encounter cases where sentences express different relations between identical argument texts. Our previous results indicate that obvious correlations between argument texts and certain relations may cause models to internalize specific shortcuts. If model decisions do not react to context changes in these cases, this causes vulnerabilities and counterintuitive decision processes.

\begin{table}
    \centering
    \caption{Evaluation on adversarial samples. We create multiple adversarial samples from each original sample. The model prediction is considered \textit{correct} for an adversarial sample if the prediction changes from the original to the adversarial sample.}
    \label{tab:adv_conf_and_performance}
    \begin{tabular}{l c c}
                    & Adv. \textit{Only Arguments}    &  Adv. \textit{Other} \\ \noalign{\smallskip} \hline \hline \noalign{\smallskip}
    \# Arg Pairs    & 12                 & 12           \\
    \# Samples      & 120                & 120          \\
    Accuracy        & .41$\pm$.31       &.84$\pm$.20    \\
    Confidence      & .88$\pm$.19       &.85$\pm$.18    \\
    \end{tabular}
\end{table}

\cite{gardnerEvaluatingModelsLocal2020} shows that deep learning models have problems reacting appropriately to specific context changes. We transfer the concept to the relation classification task introduced in \autoref{sec:task_definition}. For a given sample, we keep the argument texts and automatically change the context to change the relation between arguments with Chat GPT \footnote{\url{https://openai.com/blog/chatgpt}}. We refer to these as \textit{adversarial samples}. \autoref{tab:adv_conf_and_performance} shows how the RoBERTa base model performs on these samples. The first column refers to samples where the model predicts the final label from only the arguments with high confidence. For these samples, we suggest that models might ignore relevant context information. The second column refers to adversarial samples created from other instances. For these other instances, the model refers to the context for its decision.

The accuracy indicated the fraction of times when the adversarial adoptions lead to a changed model prediction. While the F1 score is higher on samples with semantic class \textit{Only Arguments}, the model reacts to adversarial samples less often than samples with different semantic classes. Our results indicate that deep learning models learn correlations from the dataset. These correlations may be unintuitive for humans. This causes model decisions that differ from human decision patterns for relation classification.

\subsection{Reductive Semantic Extents}
\label{sec:reductive_semantic_extents}

The previous results show how comparing extensive semantic extents for models, and humans can reveal spurious decision patterns. Determining extensive semantic extents requires a task-specific prioritization of possible candidate expansions (\autoref{sec:human_annotation}). Subsection \ref{sec:reductive_extent} shows we can interpret input reductions as reductive semantic extents. 

\begin{figure}
    \centering
    \includegraphics[scale=.5]{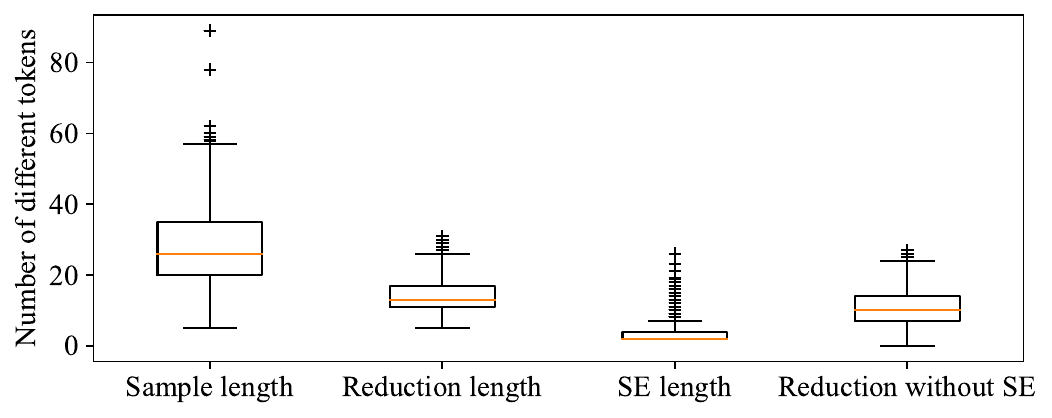}
    \caption{Average number of tokens in samples extensive and reductive (input reductions) semantic extents.}
    \label{fig:length_areas}
\end{figure}

Our previous results require identifying samples where models focus mainly on the arguments and ignore the context. In theory, input reduction methods could reduce all input tokens from such samples and create similar results compared to extensive semantic extents. \autoref{fig:length_areas} shows that reductive extents tend to contain more tokens than extensive extents. We observe hardly any cases where only the arguments remain in input reductions.

\section{Limitations}
\label{sec:limitation}

This research formalized the concept of an \textit{explanation} for relation classification. With that, we are looking at a fundamental NLP task. The concept of \textit{semantic extents} reveals latent decision patterns for this essential task. Real-world applications confront models with scenarios that are structural and conceptual more challenging. Our results demonstrate the fundamental applicability of semantic extents. Our methodology is a basis for further developments in explainable and trustworthy NLP.

We compared semantic extents for both humans and models. The comparison hints toward spurious behavioral patterns. To prove such suppositions, we applied a further analysis that includes the creation of adversarial samples. We applied Chat GPT to reduce the manual effort. The methodology presented is a tool to identify potential weaknesses. Eliminating these vulnerabilities requires a combination of semantic extents with different approaches.

Our results showed that a deep-learning model's focus on relation arguments might be too high. While our research uncovered this spurious decision pattern, there may be other spurious patterns that were not the focus of our study.  One example might be various biases. Since many data sources describe the war in Iraq from an American perspective, the classification decisions might be biased if arguments indicate that people, locations, or organizations have certain geopolitical connections.
\section{Conclusion}
\label{sec:conclusion}

The novel concept of \textit{semantic extents} formalizes explanations for the relation classification task. We defined a procedure to determine semantic extents for humans and deep learning models. A novel annotation tool supports human annotations, and a methodical framework enables the determination of semantic extents for NLP models.

We train and evaluate our models on the ACE 05 corpus. In contrast to most current applications of this dataset, we extensively use provided metadata to compare human and model decision patterns. Many previous approaches use preprocessing pipelines with no convenient access to this metadata. We implement a preprocessing pipeline on top of a uniform state-of-the-art python technology stack. 

Datasets are at risk of being affected by various biases. For example, our results suggest that particular argument texts correlate overwhelmingly with specific relation labels. NLP models recognize such patterns and base their predictions on them. If these patterns do not relate to properties of human language, model decisions become counterintuitive. Our methodology is suitable for revealing such counterintuitive ties. We want to encourage practitioners and researchers to apply our concepts or develop methodologies to increase trust in AI and NLP solutions.

Novel developments like Chat GPT draw the public's attention to NLP and AI in general. This results in new opportunities to transfer scientific innovations into practical applications. We know many weaknesses of current deep learning models and must show ways to recognize and fix them in developed approaches. Semantic extents are an essential step towards explainable information extraction systems and an important cornerstone to increasing the acceptance of AI solutions.

\bibliographystyle{splncs04}
\bibliography{master}

\end{document}